\documentclass{article}

\usepackage[final]{neurips_2023}

\setcitestyle{authoryear,open={(},close={)}}

\usepackage[utf8]{inputenc} 
\usepackage[T1]{fontenc}    
\usepackage{hyperref}       
\usepackage{url}            
\usepackage{booktabs}       
\usepackage{amsfonts}       
\usepackage{nicefrac}       
\usepackage{microtype}      
\usepackage{xcolor}         

\usepackage{graphicx}       
\usepackage{graphbox}
\usepackage{ulem}
\usepackage{lipsum}

\newif\ifcomments

\newcommand{\tag}[1]{\textless#1\textgreater}
\newcommand{\ttag}[1]{\texttt{\tag{#1}}}

\title{ICL Markup: Structuring In-Context Learning using Soft-Token Tags}

\author{%
 Marc-Etienne Brunet \\ 
  University of Toronto\\
  Vector Institute\\
  \\
  \texttt{mebrunet@cs.toronto.edu} \\
   \And
   Ashton Anderson \\
   University of Toronto\\
   Vector Institute\\ 
   \\
   \texttt{ashton@cs.toronto.edu} \\
   \And
   Richard Zemel \\
   University of Toronto \\
   Columbia University\\
   Vector Institute\\
  \texttt{zemel@cs.toronto.edu} \\
}

\begin{document}

\maketitle

\begin{abstract}
Large pretrained language models (LLMs) can be rapidly adapted to a wide variety of tasks
via a text-to-text approach, 
where the instruction and input are fed to the model in natural language.
Combined with in-context learning (ICL), 
this paradigm is impressively flexible and powerful.
However, it also burdens users with an overwhelming number of choices,
many of them arbitrary.
Inspired by markup languages like HTML, 
we contribute a method of using soft-token
tags to compose prompt templates.
This approach reduces arbitrary decisions
and streamlines the application of ICL.
Our method is a form of meta-learning for ICL;
it learns these tags in advance 
during a parameter-efficient fine-tuning ``warm-up'' process.
The tags can subsequently be used in templates for ICL on new,
unseen tasks without any additional fine-tuning. 
Our experiments with this approach yield promising initial results,
improving LLM performance on important enterprise applications such as few-shot and open-world intent detection, 
as well as text classification in news and legal domains.
\end{abstract}

\section{Introduction}
With the growing size and capabilities of large pretrained language models (LLMs),
in-context learning (ICL) has become a popular way to harness their power for new tasks.
ICL is an approach to prompting LLMs 
which includes demonstrations of how to complete the target task in the 
prompt~\citep{Dong2022ALearning}.
It has significant advantages over traditional fine-tuning, being data-efficient, highly flexible, and user-friendly. 
A LLM can be adapted to perform effectively on a new task
with only a handful of demonstrations (few-shot) and some natural language instructions.
This can be done quickly even by someone with little knowledge of machine learning. 
The LLM can also be encapsulated as a black box and shared across tasks. 
This allows individuals and organizations to leverage LLMs for new tasks,
even if they do not have the computing resources necessary to fine-tune or even host such large models.

However, ICL also has several disadvantages.
Most LLMs have not been explicitly trained or tuned to perform ICL,
and thus have not actually been optimized to approach new tasks in this format~\citep{Dong2022ALearning}. 
Like other forms of prompt engineering, 
ICL suffers from a lack of robustness across the many arbitrary choices that users encounter in the process of setting it up~\citep{Chen2022OnLearning}.
It has been shown that the performance of in-context learning can vary dramatically
based on changes to the prompt~\citep{Zhao2021CalibrateModels}.
There is also evidence to suggest that ICL performs poorly when shown 
a "none of the above" option~\citep{Kadavath2022LanguageKnow}, which could hinder its application in practical settings (e.g.\ open world classification)
where the inputs may not always correspond with any option in the label space.

\begin{figure}
    \centering
    \includegraphics[align=c, width=1.9in]{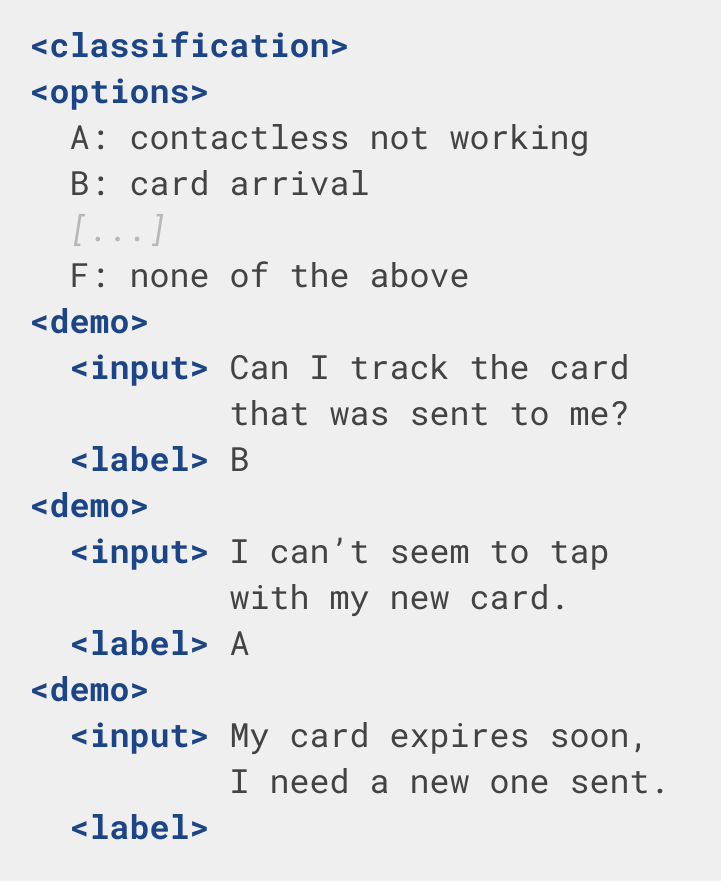}
    \hspace{0.02in}
    \includegraphics[align=c, width=2.2in]{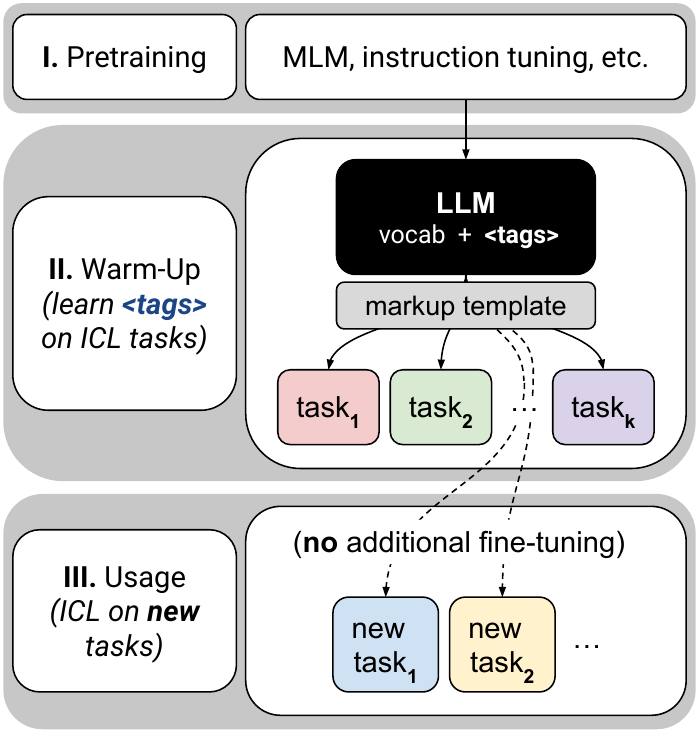}
    \includegraphics[align=c, width=1.25in]{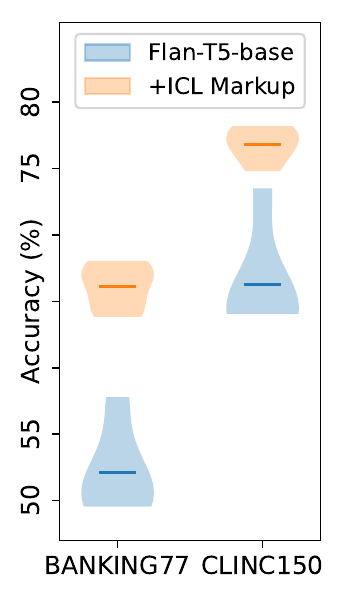}
    \caption{\textbf{Left:} Example of an ICL Markup template applied to intent detection.
        Blue boldface {\color[HTML]{1c4587}{\textbf{\textless tags\textgreater}}} indicate 
        dedicated soft-tokens introduced into the LLM's vocabulary. 
        \textbf{Center:} The soft-token tags are learned in advance 
        during a ``warm-up'' phase (parameter-efficient fine-tuning). 
        They can then be used on new, unseen tasks without additional fine-tuning. 
        \textbf{Right:} Our experiments compare hand-written prompt templates to 
        templates that utilize ICL Markup tags.
        We find that the mean accuracy (solid line in violin) on unseen tasks,
        is improved and variability (vertical extent of violin) decreases.
        This plot shows these effects on Flan-T5-base,
        where the unseen tasks are highly multi-class, 
        few-shot intent detection datasets:
        BANKING77, and CLINC150.}
    \label{fig:example-template}
\end{figure}

We propose addressing these shortcomings with an approach to ICL 
inspired by markup languages like HTML.
In this paradigm, we structure ICL prompt templates
using a dedicated set of soft-token
tags that we add to the model's vocabulary. 
These soft-tokens (a.k.a. tunable tokens) are effectively ``new words'':
bound to trainable parameters and processed like other tokens.
Their weights are learned in advance during a ``warm-up'' stage (parameter-efficient fine-tuning).
They can then be used in the ICL template for new tasks without additional fine-tuning,
and can thus also be shared across tasks.
The training process is therefore a form of meta-learning for ICL.
We show that this approach removes several arbitrary decisions from the design of ICL prompt templates.
We also provide initial empirical evidence that it can improve a model's ICL ability on new tasks.
Specifically, we show that ICL Markup improves Flan-T5 models on text classification tasks
(intent detection, news and legal domains).
We show that ICL Markup can reduce Flan-T5's 
performance variability when compared to prompt engineering,
increase classification accuracy,
as well as improve out-of-scope intent detection
when the template includes a ``none of the above'' multiple-choice option.

In our experiments on a few-shot news headline classification dataset,
we find that Flan-T5 can be very sensitive to small changes in the prompt. 
For example, switching the word used to indicate the start of a demonstration in our prompt template
from "Headline" to "Input" can impact the accuracy by up to 21 percentage points (p.p.)%
\footnote{We adopt the term "percentage points" (p.p.) to indicate an absolute 
(rather than a relative) change in percentage. 
For example, if accuracy changes from 80\% to 90\%, we would say it increased by 10 p.p.}.
A small syntactic choice like using ")" vs ":" to separate keywords from their associated values 
can impact accuracy by up to 24 p.p.,
with the direction of the effect depending on the model size.  
We also find that when compared to a search over 96 prompts,
we are able to increase Flan-T5-XL's accuracy on this dataset
from a mean of 68.9\% 
(or 70.9\% using the best prompt in our search)
to a mean of 76.8\% with ICL Markup.
This tops the previous best reported results on the dataset, 
Prompt-Based Meta-Learning ~\citep{Zhang2022Prompt-BasedClassification},
by 5.2 p.p.

\section{Background and Related Work}
\paragraph{In-Context Learning}
One can think of in-context learning (ICL) as the LLM learning the target task
``by analogy'' from the examples in the prompt~\citep{Dong2022ALearning}.
ICL has also been referred to as prompt augmentation~\citep{Liu2023Pre-trainProcessing},
since a cloze or completion-style prompt is augmented by prepending answered prompts. 
The effectiveness of ICL has sparked a great deal of research on the topic.
The areas most related to this work include pretraining methods for ICL,
for example, MetaICL~\citep{Min2021MetaICL:Context}, In-Context Tuning~\citep{Chen2022Meta-learningTuning},
and Symbol Tuning~\citep{Wei2023SymbolModels}.
Our work differentiates itself by introducing structured soft-token tags,
and performing the meta-learning in a parameter-efficient way.
ICL research also investigates methods for demonstration selection,
such as retriever systems that identify the best demonstrations
for a particular input from a pool of candidates~\citep{Liu2022WhatGPT-3}.
Other relevant work studies ICL robustness (or lack thereof) 
due to choices in prompt template designs~\citep{Zhao2021CalibrateModels}, 
or demonstration order~\citep{Lu2021FantasticallySensitivity}. 

\paragraph{Parameter-Efficient Fine-Tuning}
Prior to the rise of prompt engineering (including ICL),
the principal approach to adapting a pretrained model to a downstream task was
via fine-tuning~\citep{Liu2023Pre-trainProcessing}.
However, it is not computationally feasible for most individuals or even organizations
to fine-tune modern LLMs;
there are simply too many parameters~\citep{Ding2023Parameter-efficientModels}.
As result, several parameter-efficient approaches have been developed which enable 
fine-tuning LLMs using fewer trainable 
parameters~\citep{Lester2021TheTuning, Li2021Prefix-Tuning:Generation, Hu2021LoRA:Models}.

\paragraph{Other approaches to prompt engineering}
The work by \citet{Gu2022PPT:Learning} on pretrained prompt tuning (PPT) 
is perhaps the most related to ours.
They pretrain soft-token prompts in a self-supervised manner, 
then fine-tune them per few-shot task. 
This improves the reliablity of few-shot prompt tuning.
Our work is different because it aims to avoid the second fine-tuning step. 
Our tokens and templates are designed to be used for ICL on new tasks without further fine-tuning,
relying on demonstrations to assist with adaptation.
Prompt tuning with rules~\citep{Han2022PTR:Classification}
involves manually decomposing a task into sub-prompts,
then fine-tuning an LLM to optimize the performance of this decomposition.

\paragraph{Few-shot text classification}
The aim of few-shot text classification is to 
predict the label of a text with access to only a few annotated examples. 
We follow the common terminology of N-way K-shot,
with N indicating the number of classes, 
and K indicating the number of annotated supporting examples per class. 
These supporting examples are available for adapting the model to the task. 
Therefore, raising N and/or lowering K leads to a more challenging task.
Among the numerous few-shot text classification methods published,
Prompt-Based Meta-Learning~\citep{Zhang2022Prompt-BasedClassification}
is particularly relevant to our work
as it combines prompt tuning with meta-learning,
outperforming popular methods like Prototypical Networks~\citep{Snell2017PrototypicalLearning}, 
MAML~\citep{Finn2017Model-agnosticNetworks},
and prompt tuning~\citep{Lester2021TheTuning} on its own. 

\paragraph{Intent Detection}
An important application of text classification is intent detection,
the goal of which is to categorize the intent of a user request.
Intent detection models are used in virtual assistant (VA) and dialogue systems,
which are deployed in many enterprise use cases. 
The models in these VA systems need to be
flexible (adapting to an evolving label space),
quick to train (minute-scale),
configurable by non-machine learning experts,
capable of handling highly multi-class, few-shot, and imbalanced datasets,
and able to recognize out-of-scope intents.
A single enterprise platform provider may host over 100,000 
customer-specific models~\citep{Qian2023DistinguishAssistants}.
There has been considerable research into few-shot and open-world intent detection.
Several works have studied the use of LLMs to augment few-shot datsets 
with synthetic examples~\citep{Lin2023SelectiveV-Information, Sahu2022DataModels}.
Others explore methods to identify out-of-scope (OOS) 
examples~\citep{Khosla2022EvaluatingClassification, Zhang2020DiscriminativeInference, Qian2023DistinguishAssistants},
i.e., inputs having an intent that falls outside of the model's configured label space.  
However, \citet{Zhang2021AreDetection} note that OOS detection is 
considerably more challenging with in-domain OOS (ID-OOS) examples
which are semantically related to the in-scope intent classes.
They construct datasets in order to
examine the robustness of pretrained transformers in this challenging setting.

\section{Proposed Method: ICL Markup}
\label{sec:method}
We propose using a markup-like language to construct ICL templates
in order to reduce the number of arbitrary choices involved
and thereby enable easier and more consistent application.
Our paradigm is visualized in Figure~\ref{fig:example-template} (left).
With this approach, we separate content from presentation using soft-token tags.
For example, rather than including a demonstration in the prompt as
\texttt{statement:~I can't tap my card. class:~contactless not working},
it is included as
\texttt{\tag{input} I can't tap my card. \tag{label} contactless not working},
where \ttag{input} and \ttag{label} are soft-tokens 
that have been learned in advance to indicate the inclusion of a labeled $(X,y)$ pair. 
This removes many arbitrary choices about presentation,
e.g. whether to demarcate the demonstration using ``statement:'', ``input:'',  
``label:'' or ``category:''.
Engineers can then focus their energy on the content of the prompt,
such as the choice of demonstrations and class descriptors.
Such an approach maintains flexibility while reducing the number of arbitrary choices.

Our method is loosely inspired by markup languages like HTML or Markdown. 
When composing a webpage or a README, it has proven useful to adopt standardized syntax
in the form of special tags or tokens. 
It would be chaotic to try to interact with a browser's rendering engine using only 
natural language.
As LLMs are becoming more like general-purpose computing tools, 
it seems sensible to include standardized syntax for common tasks.
Adopting new and shared terminology has also been fruitful in
professional and academic fields.
Doing so can reduce ambiguity and avoid the communication overhead of always 
redefining concepts.
Finally, it is important to note that several such special tokens are already used by LLMs,
although they are typically hidden from users.
For example, the end-of-sentence token,~\ttag{eos},
is used to indicate the end of a textual input.

There are many possible ways to structure these markup-like tags.
Here we explore one option which targets multiple-choice style classification templates and is applicable to a broad set of tasks.
It uses the following tags:
\begin{description}
    \item[\ttag{classification}] instructs the model that the subsequent textual input is a multiple-choice classification task
    \item[\ttag{options}] demarcates the start of YAML-formatted multiple choice options;
        defines a mapping from capital letter tokens to class descriptors
    \item[\ttag{demo}] denotes the start of a labelled example
    \item[\ttag{input}] indicates the start of the example's textual input
    \item[\ttag{label}] indicates the multiple-choice letter option corresponding to the correct class descriptor
\end{description} 

We add these soft-token tags to the vocabulary of an LLM
and let the model learn them 
during a ``warm-up'' process.
This is done by composing a prompt template with the tags,
then using the template to solve ICL classification tasks
and updating the parameters associated with the tags through back-propagation.
This is a parameter-efficient fine-tuning process 
in the sense that only the parameters of the tags are updated.
The tags and template can then be used to perform ICL on new tasks without further fine-tuning.

\subsection{Learning soft-token tags}
In this work, we consider the use of ICL Markup in few-shot text classification;
only a limited number of labelled examples are available at inference for adaptation
to the target task (which is done \textit{through ICL, not parameter updates}).
We assume access to labelled data from related tasks which can be used to learn the ICL Markup tags
in advance during a "warm-up" period.
We focus on how ICL Markup can improve robustness,
and the extent to which tags learned on 
warm-up tasks can transfer value to the (unseen) target task. 

We take the following approach to warm-up.
We first construct a template using markup tokens.
This template must be appropriate for the \textit{form} of the target task,
i.e. the format of the task's inputs and outputs.
Throughout this work, the template we use is a multiple-choice style template
with a variable number of answer options,
followed by a variable number of demonstrations.
See Figure~\ref{fig:example-template} (left) as an example.
We then fine-tune the soft-tokens tags in the template using a collection of related training tasks.
We leave an exploration of using ICL Markup tags in different styles of prompt templates to future work. 

\paragraph{Implementing and initializing the soft-token tags}
An ICL Markup tag need not correspond to a single soft-token in the vocabulary,
but can be represented by an ordered set of soft-tokens (a tunable "phrase").
This increases the representational capacity that the set of tags create.
It allows a tag like \ttag{classification} to effectively replace 
the hand-written classification instructions one would prompt the model with,
and becomes the soft-prompt that adapts the model to the 
usage of the multiple-choice template.
This way, no additional soft-tokens are needed for fine-tuning;
only the parameters associated with the ICL Markup tags are updated during warm-up training.
The number of soft-tokens allocated to each tag is a hyperparameter choice.
We do not thoroughly explore the effect of this choice in this work,
but we target a \textit{total} number of soft-tokens in our prompt template between 10 and 25.
This range is based on work by~\citet{Lester2021TheTuning}, 
who examine how the performance of prompt tuning is affected 
by prompt length, i.e., the number of soft-tokens used.

There remains a question of how these soft tokens should be initialized.
In the prompt tuning literature, 
both random initialization and initialization from existing tokens in the vocabulary is 
often explored~\citep{Gu2022PPT:Learning, Lester2021TheTuning}.
We explore both in our experiments as well,
finding the choice to have a notable impact when the "none of the above" (NOTA)
option is present in the multiple-choice template.
We discuss this choice and its effects in Appendix~\ref{sec:NOTA}.
We initialize from existing tokens when the prompt template includes a NOTA option
but the task is not explicitly open-world,
otherwise we initialize randomly.

\paragraph{The relationship between training and target tasks}
The warm-up tasks used to learn the soft-token tags may differ from the target task along a few axes, e.g.,

\begin{figure}
    \centering
    \includegraphics[align=c, height=2in]{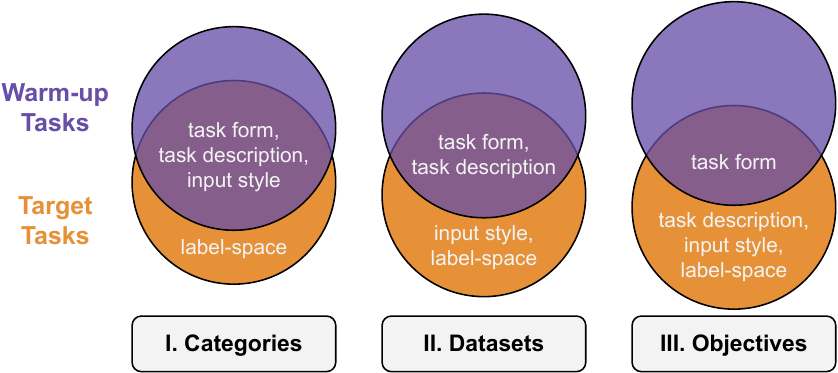}
    \caption{Relationship between the "warm-up" training tasks and the target tasks.}
    \label{fig:tasks_venn}
\end{figure}

\begin{description}
    \item[Label-space:]  The relationship between input texts and labels,
    i.e., the number classes and their definitions. 
    \item[Input style:] The characteristics of the input texts, e.g., text length, writing style;
        determined by the empirical input distribution of the dataset. 
    \item[Task objective:] The description of the objective of the task,
    in the sense of the instructions one would prompt an LLM with, 
    e.g., ``categorize these news articles'', 
    or ``determine the legal purpose of these contract provisions''.
\end{description}

We consider three types of training-target task shifts in our experiments in Section~\ref{sec:experiments}.
These are depicted in Figure~\ref{fig:tasks_venn} and described here:
\begin{description}
\item[I. Categories:] (Section~\ref{sec:huffpost}) Changes are primarily in the label space. 
    The target task consists of new categories (classes) from within the same dataset as the one used for training. 
\item[II. Datasets:] (Section~\ref{sec:icl_for_intent}) Changes in both the label space and input style.
    The target task is an entirely separate dataset, 
    with different classes and different input text characteristics,
    but sharing the same objective, e.g., intent detection.
\item[III. Objectives:] (Section~\ref{sec:ledgar}) Changes in everything except the form of the task. 
    The target task relates to the training task only in that it still involves text classification, 
    but this is on a new dataset with different objectives.
\end{description}

\section{Experiments}
\label{sec:experiments}
We use Flan-T5 models for our experiments~\citep{Raffel2019ExploringTransformer, Chung2022ScalingModels}.
We train ICL Markup tags for three model sizes: base (250M parameters), 
large (780M), and XL (3B).
Throughout all experiments,
we use the same form of multiple-choice style ICL Markup template:
having a variable number of answer options,
followed by a variable number of (labelled) demonstrations.
See Figure~\ref{fig:example-template} (left) for an example.
We use an ordered set of between 9 and 18 soft-tokens to represent the \tag{classification} tag, 
an ordered set of 2 soft-tokens for \tag{options},
and 1 soft-token for each of \tag{demo}, \tag{input}, and \tag{label}.
With the exception of our open-world evaluations, 
we use an unconstrained greedy decoding to generate the multiple choice response.%
\footnote{When the model output is not one of the multiple choice options (rarely),
we consider it to be ``none of the above'' (NOTA), or simply wrong if NOTA was not an option.}

\subsection{News headline classification (\textit{shift in categories})}
\label{sec:huffpost}
\paragraph{Datasets and setup}
We begin our experimental investigation of ICL Markup on a Huffington Post News dataset~\citep{Misra2022NewsDataset},
released for few-shot text classification by \citet{Bao2019Few-shotSignatures}.%
The dataset is partitioned along news categories (classes),
with 20 categories available for training,
5 used for validation, and the remaining 16 reserved for testing.
We first learn the ICL Markup tokens using the training categories. 
We then assess the performance on the test categories
in (5-way, 10-way) x (1-shot, 5-shot) configurations.
We use the (lowercased) class names released with the dataset to serve as the 
multiple-choice options in the prompt template,
e.g., "A: world news~~B: arts \& culture ...",
their order is randomized for each example.
When there are too many supporting examples to include all demonstrations 
in the prompt (in the 5-shot settings),
we use a maximum marginal relevance (MMR) selection strategy to retrieve the 15 most relevant
for each test instance.
This strategy is explained in greater detail in Section~\ref{sec:icl_for_intent}.
We do not enforce that every class be exemplified via a demonstration,
and we randomize the order of the demonstrations.
Further dataset details can be found in Appendix~\ref{appendix:huff_post_data}.

\begin{figure}
    \label{fig:prompt_var_decomposed}
    \centering
    \includegraphics[align=c, height=2.1in]{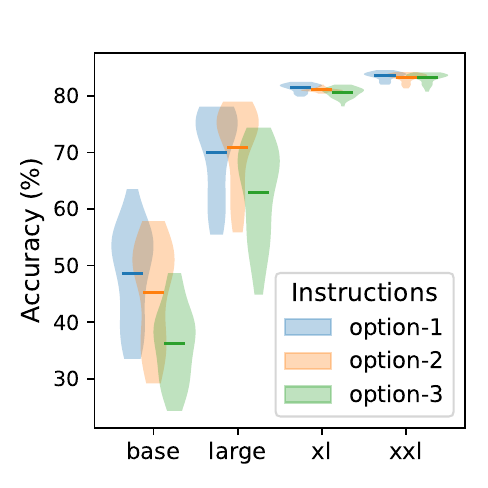}
    \hspace{0.2in}
    \includegraphics[align=c, height=2.05in]{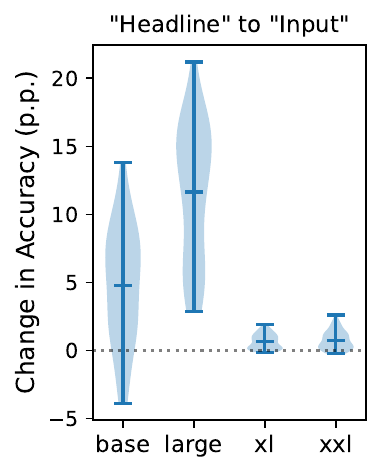}
    \hspace{-0.1in}
    \includegraphics[align=c, height=2.05in]{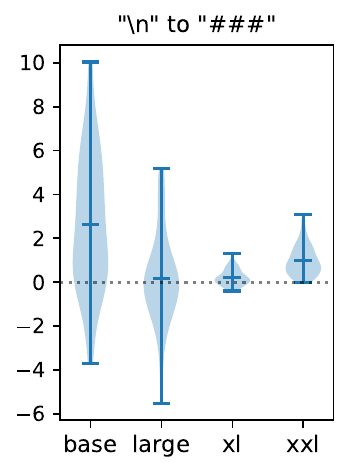}
    \caption[ICL performance variation across multiple-choice style prompts (without ICL Markup)]{\textbf{ICL performance variation across multiple-choice style prompts (without ICL Markup)}.
    Accuracy is measured via 5-way classification accuracy on the Huff Post validation set,
    averaged over 1 and 5 shot settings, and separated by model size (on the horizontal axes).
    \textbf{Left:} Empirical distribution of performance variation across three different
    options\protect\footnotemark~for the main instructions in the prompt.
    Solid horizontal lines within the violins depict the mean.
    \textbf{Right:} The effect of a single edit across otherwise unchanged prompts.
    The specific edits are indicated above each pane.
    "Headline" to "Input" is a change in the word used to indicate the start of a demonstration,
    while "$\backslash$n" to "\#\#\#" is a change in what separates the demonstrations. 
    Violins depict the empirical distribution of accuracy change (in percentage points)
    due to the edit, which is a function of the remainder of the prompt.
    When the range crosses the 0-line it is a non-monotonic edit,
    i.e., the edit may increase or decrease accuracy depending on the the prompt.}
\end{figure}

\footnotetext{The instruction options are:
    1) "Categorize the following news headlines according to their topic."
    2) "Classify these headlines based on the type of news."
    3) "Identify the type of news based on following headlines."}

\paragraph{Analyzing prompt variation}
\label{sec:huff_post_prompt_var}
In order to estimate the performance of the Flan-T5 models without ICL Markup
we conduct a prompt sweep.
We replace the tags in our multiple-choice style
ICL template with hand-engineered words and phrases,
searching over 96 different (sensible) combinations.
These are created through the cartesian product of word (or phrase) choices to replace
each of the tags presented in Section~\ref{sec:method},
roughly capturing the different axes of arbitrary decisions that must be made 
when designing a multiple-choice style prompt such as the one depicted in Figure~\ref{fig:example-template} (left).

We find a wide range (and high variance) in the accuracy
across prompts in the smaller models.
The best and worst prompt differ by 39.2 p.p. in the base model,
and 34.1 p.p. in the large model.
This gap decreases considerably in the larger XL, and XXL models,
but still amounts to 4.4 p.p. and 4.2 p.p. respectively.
This performance variance is depicted in Figure~\ref{fig:prompt_var_decomposed}.
Many small (and seemingly arbitrary) changes in the prompt can have a dramatic effect.
For example, switching only the word used to indicate the start of a demonstration in
our prompt template from "Headline" to "Input"
can impact the accuracy of Flan-T5-large by up to 21 p.p.
A small syntactic choice like using like using ")" vs ":" to separate keywords from their associated values
can impact the accuracy of Flan-T5-base by up to 24 p.p. 
The impact of this choice (and many others) depends on the model size.
For the base, large and XL models, the highest accuracy prompts use ":",
while for XXL it uses ")". 
We find several examples of non-monotonic choices within model sizes as well. 
For instance, in Flan-T5-large, switching from using a blank newline to separate demonstrations
to using "\#\#\#" can improve accuracy by 5 p.p. in some prompts,
while decreasing accuracy by 5 p.p. in others. 
This would complicate attempts to iteratively improve a prompt manually
through successive edits.
Overall, our prompt search confirms previous reports of sensitivity to arbitrary changes.

\paragraph{Results}
The results of our Huffington Post experiments are plotted in Figure~\ref{fig:huff_post_results}.
They are tabulated in greater detail in in Appendix~\ref{appendix:huff_post_extra_results}.
\textbf{We find that ICL Markup improves over the average (mean) prompt accuracy in every setting.}
It also reduces performance variation,
in the sense that the variance in accuracy across ICL Markup warm-up training 
runs that use different random seeds is lower than across different prompts.
This is especially clear in the smaller models sizes (base and large). 
With the exception of Flan-T5-large on 5-way classification,
we find that the mean accuracy of ICL Markup
improves on the \textit{best} prompt in the sweep. 
Averaged across ways/shots, ICL Markup increase Flan-T5-XL's accuracy
from a mean of 68.9\% (or 70.9\% using the best prompt in our search)
to a mean of 76.8\% with ICL Markup.
This is a considerable overall improvement.
Notably, \textbf{in every test configuration, Flan-T5-XL with ICL Markup outperforms 
Prompt-Based Meta-Learning (PBML)}~\citep{Zhang2022Prompt-BasedClassification}, 
based on the authors' reported accuracy. 
To our knowledge,
these were the best published results on this dataset%
\footnote{
Literature search conducted in October 2023.
It should be noted that \citet{Gong2023Prompt-BasedClassification} indicate that their method, PGCA,
outperforms PBML on this dataset, but they report a lower accuracy in their experiments. 
Here we compare to the \textit{highest reported} figures on this dataset,
which are attributed to PBML.},
outperforming other few-shot methods like Prototypical Networks~\citep{Snell2017PrototypicalLearning}
and MAML~\citep{Finn2017Model-agnosticNetworks}, 
as well  prompt tuning~\citep{Lester2021TheTuning}.
Furthermore, PBML relies on gradient-based parameter updates for few-shot adaption,
while ICL Markup does not.

\begin{figure}
    \centering
    \includegraphics[align=c, height=2.2in]{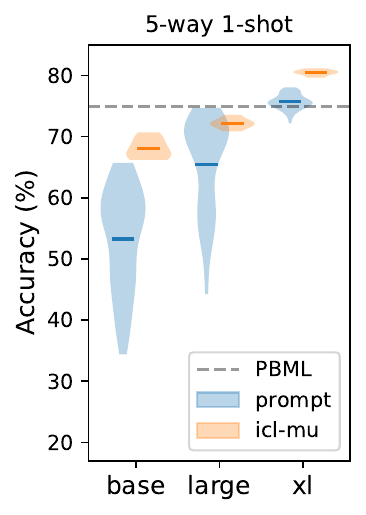}
    \includegraphics[align=c, height=2.2in]{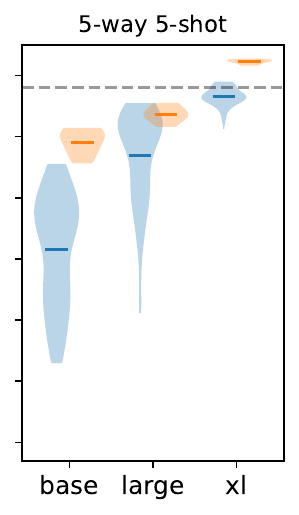}
    \includegraphics[align=c, height=2.2in]{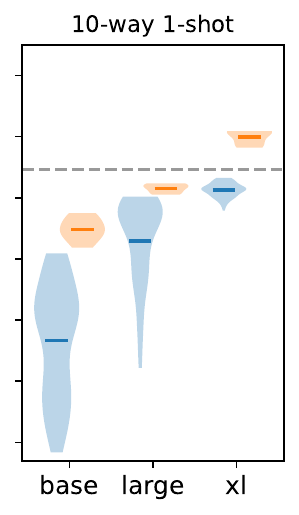}
    \includegraphics[align=c, height=2.2in]{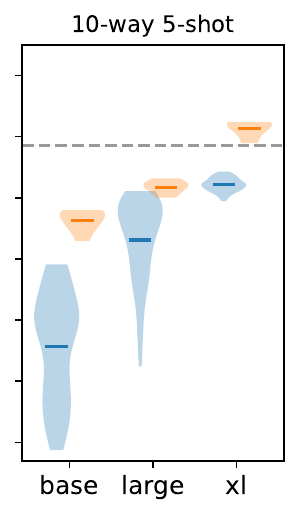}
    \caption{Few-shot Huffington Post News headline classification results over
    (5-way, 10-way) x (1-shot, 5-shot) scenarios;
    separated by model size (on the horizontal axes).
    The blue violins show the empirical distribution of test accuracy across the 96
    prompt templates searched. The orange violins show the distribution of test accuracy
    across 5 independent ICL Markup training runs. 
    The solid line within each violin indicates the mean.
    The previous best reported accuracy, PBML, is shown with the gray dashed line.}
    \label{fig:huff_post_results}
\end{figure}

\subsection{Intent detection (\textit{shift in datasets})}
\label{sec:icl_for_intent}
\begin{figure}[b]
    \centering
    \includegraphics[align=c, width=\textwidth]{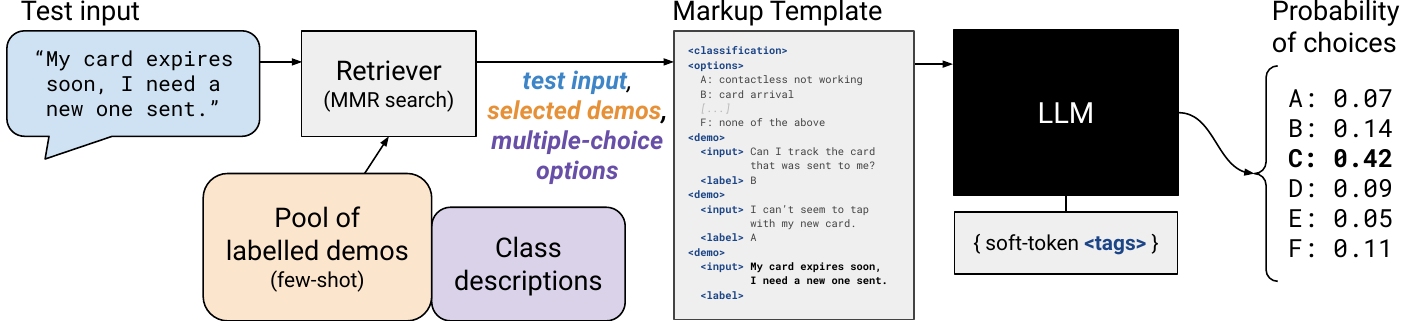}
    \caption{Our pipeline for using ICL Markup in intent detection.}
    \label{fig:pipeline}
\end{figure}

We now consider ICL Markup in few-shot and open-world intent detection tasks.
This is an important practical setting that stands to benefit from the flexibility of ICL.
Intent categories often change, and data is limited. 
However, intent detection can be highly multi-class.
If the LLM's context window is limited,
there may not be enough space in the context window
to include a demonstration for each intent class.
In extreme cases, i.e., 1000+ intent classes,
there may not even be enough context space to enumerate all of the intents.
We address this by retrieving the k most relevant demonstrations for an input instance 
from a (few-shot) candidate pool, similar to~\citet{Liu2022WhatGPT-3}.
However, rather than selecting the k-nearest-neighbors,
we use a maximum marginal relevance (MMR) selection strategy~\citep{Carbonell1998TheSummaries},
encouraging diverse demonstrations as proposed by~\citep{Ye2022ComplementaryLearning}.
We use the set of labels of the k-selected demonstrations 
to re-scope the classification for each test instance,
narrowing the label space (per test instance) to something that fits in context.
This MMR retriever approach requires little overhead beyond a vector database 
and a lightweight embedding model.
Figure~\ref{fig:pipeline} illustrates this retriever-controlled ICL Markup pipeline. 

\subsubsection{Few-shot evaluation}
\label{sec:fsid_eval}
\paragraph{Datasets and setup}
We train and test using the few-shot datasets released by~\citep{Zhang2021AreDetection}.
In order to learn the weights for the soft-token tags we build a training set 
from four of the intent detection datasets.
We include HWU64, SNIPS, and ATIS. 
We also include either BANKING77 or CLINC150 and keep the other
as the target task for testing.
With this approach, the target task dataset is withheld during 
the soft-token learning phase,
and thus remains entirely unseen until inference.
Further dataset details are described in Appendix~\ref{appendix:datasets}.
In both training and testing, 
we use the MMR retriever to select k-demonstrations per input instance
\footnote{We use $k=9$ for the intent detection tasks} 
\footnote{We use SBERT \texttt{all-mpnet-base-v2} embeddings\citep{Reimers2019Sentence-BERT:BERT-Networks}}.
We trim the set of demonstrations for a test instance
if they do not all fit in the model's context window.
With this setup, the true label appears among the multiple choice options in 
approximately 97\% of the examples in the training set.
To address the 3\% of examples without a correct multiple-choice answer,
we include a ``none of the above'' option in the prompt template used for intent detection.

At test time, we evaluate on BANKING77 or CLINC150 (whichever was withheld when the tags were learned).
We use the fixed 5-shot or 10-shot training splits released by~\citet{Zhang2020DiscriminativeInference}%
\footnote{At the time of writing datasets were available at \url{github.com/jianguoz/Few-Shot-Intent-Detection}}.
These training splits serve as the demonstration pool for the MMR retriever.
The model is thus adapted to the test (target) task using only ICL. 

\paragraph{Baselines}
To estimate the ICL performance of the Flan-T5 model on these tasks (without our markup tags),
we replace the tags in our ICL template with hand-engineered words and phrases,
but otherwise use the same pipeline (Figure~\ref{fig:pipeline}).
We focus on Flan-T5 XL for these experiments, 
though we do consider the base sized model as well.
We consider 5 sets of (sensible) words and phrases.
These choices are listed in Appendix~\ref{appendix:templates}.
We take this approach to isolate the effect of using ICL Markup
from the effect of using the MMR retriever.
We report the mean and standard deviation across the different prompts.

We also compare to the best reported results that we could find on these
datasets\footnote{Literature searches were conducted in July 2023.}:
In-Context Data-Augmentation~\citep{Lin2023SelectiveV-Information},
as well as the best reported prompt-based baseline:
Prompt Tuning with Rules (PTR)~\citep{Han2022PTR:Classification}.
We note that both ICDA and PTR fine-tune models on the target task data,
while ICDA additionally uses synthetic data augmentation,
and PTR involves hand-crafting rule-based prompts from the class names.
Thus both methods require considerable effort to configure for a target task. 
By contrast, ICL Markup keeps the model fixed,
requiring only sensible class names and test-time access to the few-shot demonstration pools. 
Finally, we include a comparison with ChatGPT.
First in a naive configuration:
where all label-options are listed and 9 random demonstrations are included for each test instance.
Then we compare to using ChatGPT in the MMR retriever pipeline (Figure~\ref{fig:pipeline}),
as a drop-in replacement for Flan-T5.

\begin{table}
\centering
\caption{Few-shot intent detection accuracy (mean $\pm$ std-dev.)}
\label{tab:results-few-shot}
\begin{tabular}{lccllll}
{} & fine&  data& \multicolumn{2}{c}{BANKING77} & \multicolumn{2}{c}{CLINC150} \\
{} & tune& aug.& 10-shot & 5-shot & 10-shot & 5-shot \\
\toprule
ICDA~\citep{Lin2023SelectiveV-Information} & yes & yes & 89.8 & 84.0 & 94.8 & 92.6 \\
PTR~\citep{Han2022PTR:Classification} & yes & no & 86.0 & 79.7 & 90.8 & 90.2 \\
ChatGPT (naive) & no & no & 70.9 & - & 86.5 & - \\
ChatGPT (MMR) & no & no & 79.6 ± 0.7 & 74.6 ± 1.0 & 86.7 ± 1.1 & 84.3 ± 1.5 \\
\midrule
Flan-T5-XL (MMR) & no & no & 82.3 ± 0.5 & 78.5 ± 0.7 & 89.6 ± 0.2 & 87.9 ± 0.2 \\
~~+ICL-MU & no & no & 85.5 ± 0.6 & 82.1 ± 0.4 & 91.0 ± 0.1 & 88.8 ± 0.1 \\
 \bottomrule
\end{tabular}
\end{table}

\paragraph{Results}
The results of our few-shot evaluation are presented in Table~\ref{tab:results-few-shot}.
\textbf{We find that ICL Markup improves the Flan-T5-XL model in all cases}, 
pushing its mean performance (over datasets/shots) to be on par with PTR, 
which fine-tunes on target task data. 
We found the gains from using ICL Markup to be even greater in the smaller base model.
For example,
the performance of ICL Markup with Flan-T5-base on CLINC150 and BANKING77 (10-shot)
is shown in Figure~\ref{fig:example-template} (right),
where we see a clear and substantial improvement. 
Somewhat surprisingly,
Flan-T5-XL outperforms both of the ChatGPT baselines we tried 
(even without markup).

\subsubsection{Open-world (few-shot) evaluation}
Next we consider few-shot \textit{open-world} intent detection.
Open-world classification involves the identification of out-of-scope (OOS) inputs, 
requiring a model to identify inputs that do not belong to any of the predetermined classes.

\paragraph{Datasets and setup}
We evaluate on the open-world variants of BANKING77 and CLINC150 released by~\citep{Zhang2021AreDetection}:
BANKING77-OOS,
CLINC-Single-Domain-OOS-banking,
and CLINC-Single-Domain-OOS-credit-cards.
These datasets are particularly challenging because they are designed to contain 
\textit{in-domain} out-of-scope (ID-OOS) examples which are semantically similar to the 
in-scope intent classes.
They also contain out-of-domain out-of-scope (OOD-OOS) examples,
which are easier to recognize as out of scope. 

We use the same soft-token tags that were learned for the few-shot (closed-world) experiements (Section~\ref{sec:fsid_eval}).
We use the tags learned without BANKING77 to evaluate on BANKING77-OOS,
and the tags learned without CLINC150 to evaluate on the CLINC OOS variants. 
The composition of these target task datasets is such that
they were completely unseen while the corresponding soft-token tags were learned. 

Recall that the tags for intent detection were learned in a prompt template that 
always included a ``none of the above'' (NOTA) multiple-choice option.
For these open-world evalutaions we consider a NOTA multiple-choice selection to be a prediction of OOS.
It has been shown that large LLMs can be negatively affected 
both in accuracy and calibration when they are presented with a ``none of the above''
option~\citep{Kadavath2022LanguageKnow}.
We explore whether exposure to such an option during the warm-up 
phase may allow the LLM to select it with more success. 

\paragraph{Baselines and OOS thresholds}
We again compare to Flan-T5 without markup. 
For our open-world evaluation,
we first choose the best prompt on a warm-up task validation set.
This is selected from among the prompt options explored in the previous 
few-shot (closed-world) evaluation.
We then report the performance of Flan-T5 on that prompt.

We also compare to the baselines released with the OOS datasets. 
For the open-world evaluation we follow a procedure very similar to the one used to produce these
baselines.
We consider ID-OOS and OOD-OOS examples separately, 
and average our results over ten random 5-shot or 10-shot draws from each 
of the target task's training sets.
However, \textit{unlike the baselines,
we do not use target task validation data to select the OOD-OOS thresholds}
for Flan-T5 or Flan-T5 with our markup.
When assessing against OOD-OOS examples,
we simply use the model's (greedy) generated prediction,
taking ``none of the above'' to correspond to OOS. 
However, like the previous baselines, 
when assessing against the challenging ID-OOS examples,
we increase the OOS sensitivity by tuning a threshold using the target task's validation set.
If the model has assigned a probability above this threshold to the  
``none of the above'' multiple choice option, 
we interpret the prediction as OOS. 

\begin{figure}
    \centering
    \includegraphics[align=c, height=2.5in]{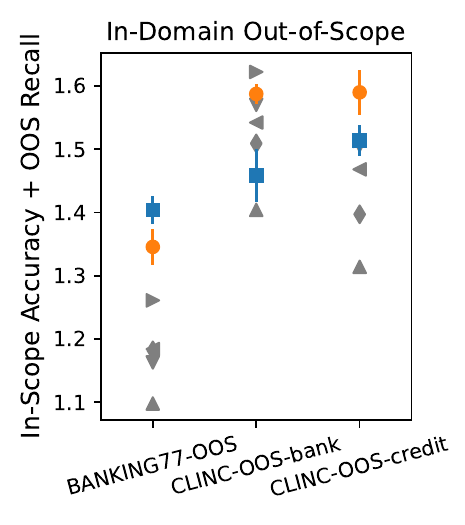}
    \hspace{0.05in}
    \includegraphics[align=c, height=2.5in]{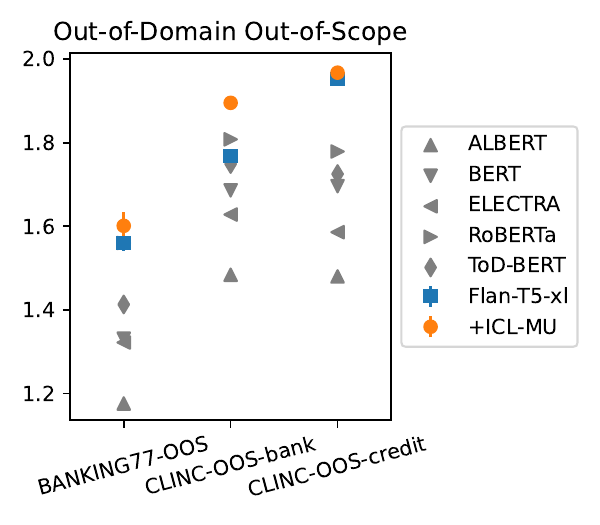}
    \caption{\textbf{Evaluation on open-world (5-shot) intent detection tasks}.
    The baselines by \citet{Zhang2021AreDetection} (gray triangles/dimonds)
    are fine-tuned per task (dataset) on the training set,
    then use the validation set to tune separate ID-OOS and OOD-OOS thresholds,
    aiming to maximize accuracy on in-scope intent classes + OOS recall. 
    In contrast, the Flan-T5 baseline and ICL Markup (ICL-MU) do not fine-tune parameters 
    (or the OOD-OOS threshold) using target task data.
    Points are means over ten 5-shot draws;  Flan-T5 and ICL-MU error bars are $\pm$std-dev.}
    \label{fig:open-world-5-shot}
\end{figure}

\paragraph{Results}
The results of the 5-shot open world experiment are shown in Figure~\ref{fig:open-world-5-shot}.
The 5 baselines are from work by \citet{Zhang2021AreDetection}.
They were optimized to maximize the in-scope accuracy plus out-of-scope recall.
Separate models are trained for each dataset-shot combination,
and OOD-OOS and ID-OOS thresholds were tuned separately on the validation sets.
\textbf{ICL Markup outperforms all previous baselines in 5 of the 6 5-shot configurations} (7 of 12 total).
It also outperforms our Flan-T5-XL baseline in 5 of the 6 5-shot configurations (10 of 12 total).
While our markup does reduce performance on BANKING77 ID-OOS compared to the baseline Flan-T5-XL,
the overall (mean) change across the 12 configurations is positive in favor of ICL Markup. 

See Appendix~\ref{appendix:open-world} for complete results.

\begin{table}[b]
    \centering
    \caption{Accuracy on LEDGAR legal text classification dataset}
    \label{tab:ledgar}
    \begin{tabular}{lcl}
    \toprule
    MMR Retriever & 19.1 & (guess on multiple-choice) \\
    1-NN & 80.1 & (using SBert embeddings) \\
    \midrule
    Flan-T5-XL & 79.2 $\pm$ 1.6 &\\
    +ICL-MU & 81.7 $\pm$ 0.4 & (p-value 0.024)\\
    \bottomrule
    \end{tabular}
\end{table}

\subsection{Legal text classification (\textit{shift in objectives})}
\label{sec:ledgar}
It is also interesting to understand whether ICL Markup tags learned on one task objective
can add value when attempting a different objective.
We conduct a limited experiment to probe whether this occurs.
We evaluate the ICL Markup tags learned for few-shot intent detection 
(Section~\ref{sec:fsid_eval})
on a version of LEDGAR~\citep{Tuggener2020LEDGAR:Contracts},
a 100-way text classification dataset in the legal domain.
We follow the same approach as for our few-shot evaluation in Section~\ref{sec:icl_for_intent},
setting k=7 (since the input texts are longer).
However, this task is not few-shot; 
all training examples are available to the MMR retriever
as candidate demonstrations.
The results are presented in Table~\ref{tab:ledgar}.
While the results do not improve much beyond a nearest-neighbour baseline,
they nonetheless show that the soft token tags,
which were \textit{trained on only intent detection tasks},
can nonetheless help the model improve its ICL ability
with this unrelated task.

\section{Discussion}
In-context learning (ICL) offers great flexibility for application-developers
wishing to leverage LLMs, 
but it also lacks robustness and leads to arbitrary decisions which may 
significantly affect the system's performance.
Our markup-inspired proposal offers structure to help minimize these situations,
and our experimental results are promising.
When compared to hand-crafted prompts, 
ICL Markup can reduce variability and improve performance.
We believe this is especially interesting in an application area like intent detection
which stands to gain from the increased flexibility of ICL.
Our few-shot Huffington Post classification results,
and open-world intent detection results are noteworthy in and of themselves.
They indicate that the benchmarks on these datasets can be 
matched or outperformed using ICL for few-shot adaptation,
rather than relying on parameter updates.
The ICL performance can then be further improved by using our markup templates.

\subsection{Limitations \& future work}
Our experiments are limited, especially in the models examined. 
The LLMs considered (encoder-decoders of 250M, 780M and 3B parameters) are small compared to the state of the art.
Additionally, our experimental scope is limited to classification tasks,
but ICL has much broader applications.
We see an obvious next step being the scaling up of experiments to more
diverse LLM architectures and larger sizes.
It would be especially interesting to expand on the \textit{shift in objectives} setting,
building a very diverse pool of warm-up tasks to produce 
general purpose tags. 
In this setting, one could explore composing tokens into different prompt templates,
and adapting to tasks beyond classification. 
If the warm-up task pool was sufficiently large and diverse,
it may also be interesting to try fine-tuning all the model parameters
along with the tags.
Finally, we think it could be interesting to introduce a "none of the above" tag, \ttag{nota},
to capture the idea of an answer being out-of-scope.
This could be included as an explicit multiple-choice (MC) option, 
or the model could be encouraged to generate this tag (instead of the MC options)
whenever the scope of possible answers presented does not match the question. 

\begin{ack}
Resources used in preparing this research were provided, in part, 
by the Province of Ontario, the Government of Canada through NSERC and CIFAR,
and companies sponsoring the Vector Institute\footnote{\url{https://vectorinstitute.ai}}.
We would like to thank the University of Toronto computing and administrative staff for their support,
the anonymous reviewers for their insightful feedback, 
and Tom Zollo, Zhun Deng, and Lillio Mok for their helpful comments. 
\end{ack}

\newpage
\bibliography{references}


\newpage

\appendix
\section{Effect of initialization and NOTA selection}
\label{sec:NOTA}
In the prompt tuning (and tunable token) literature, 
both random initialization and initialization from existing tokens in the vocabulary is often 
explored~\citep{Gu2022PPT:Learning, Lester2021TheTuning}.
We explore both in our experiments as well.
We try initializing tokens from random parameter values,
as well as from existing tokend in the vocabulary.
We refer to the latter strategy as "anneal".
We find that this choice to has a notable impact 
when the "none of the above" (NOTA) option is present
in the multiple-choice template.

In the Huffington Post experiments, 
all possible classes are included in the prompt as multiple choice options. 
We initialize the tokens randomly.
We explored the anneal strategy as well, but did not see a notable difference in behavior.

In the intent detection experiments, 
there are too many classes to include all as multiple choice options.
We limit the options using the MMR retriever
as discussed in Section~\ref{sec:icl_for_intent}.
This process does not guarantee the inclusion of the correct class among the options.
We therefore include "none of the above" (NOTA) as the final multiple choice option 
in template.
This leads to two notions of accuracy:
\textit{multiple-choice (MC) accuracy}, 
which considers whether the LLM answers the multiple-choice questions correctly,
and \textit{task accuracy}, which further requires the MMR retriever to have included 
the true label among the multiple-choice options. 
Mathematically, task accuracy must be less than or equal to MC accuracy. 
We report task accuracy throughout the main text. 

We find that the anneal strategy leads to better task accuracy,
while the random initialization leads to better NOTA usage.
See Table~\ref{tab:nota}.
Therefore, for intent detection,
where a NOTA option is present in the ICL Markup template,
we use the annealed tokens for the few-shot (closed world) classification.
LEDGAR also has a NOTA option in the template, 
so we evaluate using the annealed tokens.
For the open-world classification,
where part of the task objective is to distinguish between in-scope and out-of-scope intents,
we use the random-initialized tokens.
We additionally find these tokens to be more robust while tuning OOS thresholds.

\begin{table}[b]
\centering
\caption{Effect of initialization on "none of the above" (NOTA) usage}
\label{tab:nota}
\resizebox{\textwidth}{!}{%
\begin{tabular}{lllrrlrrrrr}
 & & & \multicolumn{2}{c}{Accuracy}&  &\multicolumn{5}{c}{None of the Above}\\
\toprule
dataset& shot& model& task& MC&  ~~& actual & pred.& recall& prec.& F1\\
\midrule
BANKING77 & 5& Flan-T5-XL & 78.5 & 78.9 &  &6.0& 1.6 & 6.6 & 29.2 & 9.0 \\
& & +ICL-MU (rand) & 81.1 & 82.2 &  && 4.4 & 18.5 & 25.3 & \textbf{21.2} \\
& & +ICL-MU (anneal) & \textbf{82.1} & 82.3 &  && 0.6 & 3.3 & 36.0 & 6.0 \\
\midrule
& 10& Flan-T5-XL & 82.3 & 82.5 &  &4.1 & 1.2 & 5.2 & 18.3 & 6.9 \\
& & +ICL-MU (rand) & 84.9 & 85.6 &  && 3.1 & 17.5 & 23.5 & \textbf{19.8} \\
& & +ICL-MU (anneal) & \textbf{85.5} & 85.6 &  && 0.4 & 2.1 & 19.8 & 3.8 \\
\midrule
CLINC150 & 5& Flan-T5-XL & 87.9 & 88.4 &  &4.6 & 0.7 & 9.4 & 68.7 & 15.4 \\
& & +ICL-MU (rand) & 88.3 & 88.9 &  && 1.4 & 12.2 & 39.5 & \textbf{18.3} \\
& & +ICL-MU (anneal) & \textbf{88.8} & 89.1 &  && 0.4 & 6.3 & 70.8 & 11.5 \\
\midrule
& 10& Flan-T5-XL & 89.6 & 89.8 &  &3.4 & 0.5 & 7.7 & 50.8 & 12.1 \\
& & +ICL-MU (rand) & 90.5 & 90.9 &  && 1.1 & 12.0 & 36.9 & \textbf{17.7} \\
& & +ICL-MU (anneal) & \textbf{91.0} & 91.3 &  && 0.5 & 7.2 & 50.9 & 12.6 \\
\bottomrule
\end{tabular}
}
\end{table}

\subsection{Tuning OOS Thresholds}
\begin{figure}[!h]
    \centering
    \includegraphics[align=c, width=2.5in]{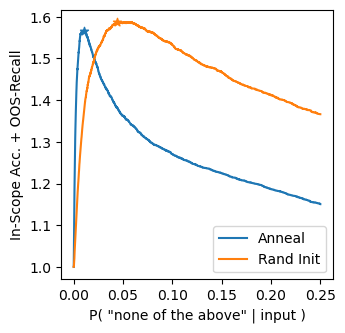}
    \caption{ID-OOS threshold sweep on validation set}
    \label{fig:threshold_sweep}
\end{figure}

As mentioned in Section~\ref{sec:experiments},
when assessing against out-of-domain out-of-scope (OOD-OOS) examples,
we simply use the model's (greedy) generated prediction,
taking ``none of the above'' to correspond to OOS. 
However, when assessing against the challenging ID-OOS examples,
we increase the OOS sensitivity by tuning a threshold using the target task's validation set.
If the model has assigned a probability above this threshold to the  
``none of the above'' multiple choice option, 
we interpret the prediction as OOS. 
In Figure~\ref{fig:threshold_sweep} we show the performance of ICL Markup
vs. different threshold values.
We notice that initializing tokens randomly leads to more robustness with regards 
to threshold choice. 
There is a flatter optima that is more tolerant to 
a small mismatch between the test and validation set. 
We use this initialization strategy for the open-world evaluation. 

\section{Huffington Post Experiments}

\subsection{Dataset details}
\label{appendix:huff_post_data}
We experiment using a Huffington Post News dataset~\citep{Misra2022NewsDataset}
processed and released for meta-learning and few-shot learning by~\citet{Bao2019Few-shotSignatures}%
\footnote{The few-shot dataset is released with pretokenized text.
Rather that use it in this way, 
we map the tokenized headlines back to the original dataset and use the original headlines instead,
preserving capitalization and punctuation}.
The goal is to classify news headlines into their corresponding news category,
e.g. ``World News'', ``Arts \& Culture''.
There are 41 such categories that have been divided up in 20 categories for training,
5 for validation, and 16 for testing.
We build training, validation, and test sets using a data loader released 
by~\citet{Zhang2022Prompt-BasedClassification}.
The data loader samples random few-shot episodes, 
first choosing 5 or 10 classes (5-way or 10-way),
then choosing 1 query example for each class,
then choosing 1 or 5 supporting examples per class available as demonstrations for ICL (1-shot or 5-shot).
Each of our test sets is composed of 10,000 query examples
(which is 2000 episodes in 5-way testing, and 1000 episodes in 10-way testing).
The baseline Flan-T5 and Flan-T5+ICL-MU models are tested on the same query examples. 
We learn the ICL Markup tokens using only the training categories,
sampling 5000 episodes from each of the (5-way, 10-way) by (1-shot, 5-shot) configurations.
We use this combined training set for all ICL-Markup training runs, 
i.e., our tokens are not way/shot specific.

\begin{table}[!h]
    \centering
    \caption{HuffPost dataset composition}
    \begin{tabular}{ccccc}
         \toprule
         Text Length (avg.)&  Example/Class&  Train Classes&  Val. Classes& Test Classes\\
         \midrule
         11&  900&  20&  5& 16\\
         \bottomrule
    \end{tabular}
    \label{tab:huff_post_data}
\end{table}

\subsection{Tabulated results}
\label{appendix:huff_post_extra_results}

\begin{table}[!h]
\centering
\caption{Few-shot classification accuracy on Huffington Post dataset. Mean ± stddev (best on val.)}
\begin{tabular}{lcccc}
\toprule
& \multicolumn{2}{c}{\textbf{5-way}} & \multicolumn{2}{c}{\textbf{10-way}} \\
 & 1-shot & 5-shot & 1-shot & 5-shot \\
\midrule
PBML~(\citeauthor{Zhang2022Prompt-BasedClassification}) & 74.9 & 78.0 & 64.6 & 68.6 \\
\midrule
Flan-T5-base & 53.3±8.4 (64.5) & 51.5±9.1 (63.4) & 36.7±9.3 (50.6) & 35.7±8.8 (49.0) \\
~~+ICL-MU & 67.8±1.2 (69.7) & 68.9±1.9 (70.9) & 54.6±1.6 (56.7) & 56.1±1.8 (57.9)\\
\midrule
Flan-T5-large & 65.4±7.8 (74.7) & 66.9±7.6 (75.0) & 52.9±7.1 (59.4) & 53.1±6.9 (59.7) \\
~~+ICL-MU & 72.1±0.9 (73.6) & 73.3±1.0 (74.7) & 61.5±0.8 (62.3) & 61.5±0.9 (62.4) \\
\midrule
Flan-T5-XL & 75.7±1.2 (77.8) & 76.5±1.5 (78.7) & 61.3±1.2 (63.3) & 62.1±1.1 (63.9) \\
~~+ICL-MU & 80.4±0.6 (\textbf{81.2}) & 82.5±0.2 (\textbf{82.7}) & 70.3±0.9 (\textbf{70.9}) & 71.8±0.7 (\textbf{72.4}) \\
\bottomrule
\end{tabular}
\label{tab:huffpost}
\end{table}

For each model size,
we report the mean performance of the prompt sweep $\pm$ 1 standard deviation. 
In parentheses, we report the test performance of the best prompt as
determined with the validation set. 
Below each, we report the mean and standard deviation of ICL Markup (ICL-MU), 
taken over several training runs with different random seeds.
In parentheses, we report the test performance of the best training run as
determined with the validation set.

\newpage
\section{Intent Detection Experiements}
\subsection{Training set composition}
\label{appendix:datasets}
In order to learn the weights for the soft-token tags we built a training set 
from three intent detection datasets:
HWU64, SNIPS, and ATIS. 
We further include either BANKING77 or CLINC150,
the one that is \textit{not} the target for testing. 
These datasets are described in Table~\ref{tab:test-sets}.
Specifically, ICL Markup tags tested on BANKING77 and BANKING77-OSS
are trained with dataset splits marked with an `B',
and hyper-parameters and the ID-OOS threshold are tuned with those marked with a `b'.
Tokens tested on CLINC150 and CLINC-Single-Domain-OOS
are developed in the same way with the splits marked `C' and `c'.
Dataset splits that have been crossed out were unused 
by our ICL Markup models.
During the ``warm-up'' process where the tokens are learned,
we draw on examples from the training sets to serve as inputs,
and use the validation sets to serve as the demonstration pools. 
Note there are two CLINC-Single-Domain-OOS datasets: 
banking and credit-cards.
They have the same number of intents classes
and the same number of instances in each split.
Also note that only few-shot sub-samples of the training sets were used
for hyperparameter tuning and threshold choices.

\begin{table}[!h]
    \centering
\caption{Intent detection datasets}
\label{tab:test-sets}
    \begin{tabular}{lcccc} 
         &  Intents&  Train&  Valid& Test\\
\toprule
 CLINC150&  150&  15,000\textsuperscript{B} &  3,000\textsuperscript{B} & 4,500\textsuperscript{b}\\ 
 BANKING77& 77& 8,622\textsuperscript{C}& 1,540\textsuperscript{C} & 3,080\textsuperscript{c}\\ 
 HWU64& 64& 8,954\textsuperscript{B,C}& 1,076\textsuperscript{B,C} & \sout{1,076}\\ 
 SNIPS& 7& 13,084\textsuperscript{B,C}& 700\textsuperscript{B,C} & \sout{700}\\ 
 ATIS& 17& 4478\textsuperscript{B,C}& 500\textsuperscript{B,C} & \sout{893}\\ 
\midrule
 CLINC-SD-OOS (x2)&  10&  500\textsuperscript{c} &  500\textsuperscript{c} & 500 \\ 
 ~~~~~~id-oos& & -& 400\textsuperscript{c} & 350 \\ 
 ~~~~~~ood-oos& & -& \sout{200} & 1,000 \\ 
 BANKING77-OOS&  50&  5,905\textsuperscript{b} & 1,506\textsuperscript{b} & 2,000 \\ 
 ~~~~~~id-oos& & -& 530\textsuperscript{b} & 1,080 \\ 
 ~~~~~~ood-oos& & -& \sout{200} & 1,000 \\ 
    \end{tabular}
\end{table}

\subsection{Intent detection dataset preprocessing}
All intent detection datasets are lower-cased. 
For each dataset, the descriptive class names
(which form the multiple-choice options in the filled prompt template)
are derived from the included label names.
The processing is very simple, limited mostly to changing underscores to spaces. 
The exceptions were:
\begin{itemize}
    \item removing the substring ``atis\_'' preceeding all ATIS labels
    \item changing ``flight no'' to ``flight number'' (also ATIS)
    \item separating the words in the seven SNIPS labels, 
    e.g., ``addtoplaylist'' becomes ``add to play list''
\end{itemize}

\newpage
\subsection{Additional open-world results}
\label{appendix:open-world}
\begin{figure}[!h]
    \centering
    \includegraphics[align=c, height=2.5in]{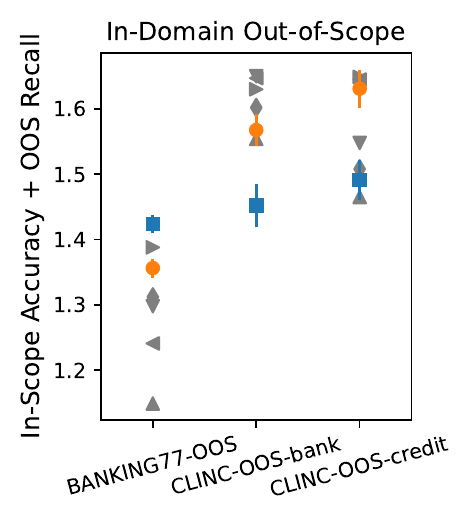}
    \hspace{0.05in}
    \includegraphics[align=c, height=2.5in]{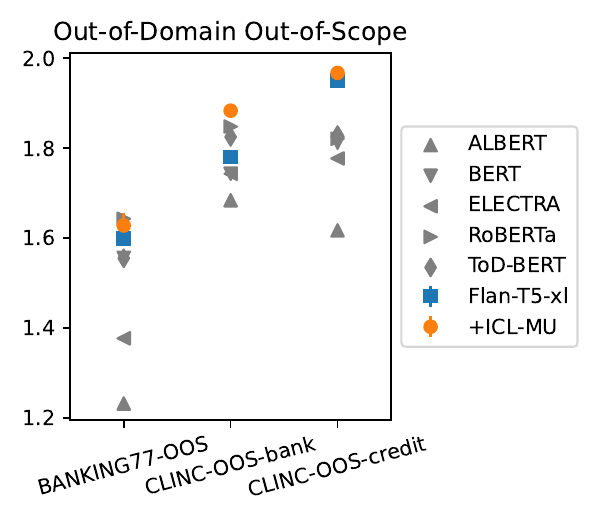}
    \caption{Evaluation on open-world (10-shot) intent detection tasks.
    Baselines~\citep{Zhang2021AreDetection} (gray triangles/dimonds)
    are fine-tuned per task (dataset) on the training set,
    then use the validation set to tune separate ID-OOS and OOD-OOS thresholds,
    aiming to maximize accuracy on in-scope intent classes + OOS recall. 
    In contrast, the Flan-T5 baseline and ICL Markup (ICL-MU) do not fine-tune parameters 
    (or OOD-OOS thresholds) using target task data.
    Points are means over ten 10-shot draws;  Flan-T5 and ICL-MU error bars are $\pm$std-dev.}
    \label{fig:open-world-10-shot}
\end{figure}

\begin{table*}[!h]
\centering
\resizebox{\textwidth}{!}{%
\begin{tabular}{lllllllllllllll}
\toprule
& & data\_source & \multicolumn{4}{l}{BANKING77-OOS} & \multicolumn{4}{l}{CLINC-OOS-banking} & \multicolumn{4}{l}{CLINC-OOS-credit-cards} \\
& & {} & & IS-Acc & OOS-Rcl & OOS-Prc & & IS-Acc & OOS-Rcl & OOS-Prc & & IS-Acc & OOS-Rcl & OOS-Prc \\
\midrule
5-shot & id-oos & ALBERT & & 20.3 & 89.5 & 39.8 & & 54.1 & 86.3 & 57.9 & & 55.5 & 75.9 & 55.8 \\
& & BERT & & 25.4 & 90.9 & 41.3 & & 75.2 & 81.8 & 70.8 & & 74.1 & 76.5 & 68.1 \\
& & ELECTRA & & 30.9 & 87.5 & 43.0 & & 64.8 & 89.4 & 65.1 & & 71.0 & 75.8 & 67.1 \\
& & RoBERTa & & 43.0 & 83.1 & 46.3 & * & 83.8 & 78.4 & 78.6 & & 64.5 & 86.8 & 63.3 \\
& & ToD-BERT & & 35.5 & 82.7 & 43.8 & & 75.1 & 75.8 & 69.4 & & 67.4 & 72.3 & 61.3 \\
& & Flan-T5-xl & * & 59.0 & 81.4 & 61.1 & & 68.4 & 77.4 & 67.3 & & 74.9 & 76.5 & 68.3 \\
& & +ICL-MU & & 56.6 & 77.9 & 56.6 & & 78.7 & 80.0 & 74.7 & * & 82.1 & 76.9 & 75.4 \\
\midrule
& ood-oos & ALBERT & & 20.3 & 97.3 & 39.9 & & 63.1 & 85.3 & 83.4 & & 55.5 & 92.5 & 81.5 \\
& & BERT & & 39.0 & 94.1 & 49.0 & & 75.2 & 93.4 & 88.8 & & 74.1 & 95.5 & 88.4 \\
& & ELECTRA & & 39.1 & 93.1 & 48.7 & & 75.5 & 87.3 & 88.8 & & 71.0 & 87.6 & 87.0 \\
& & RoBERTa & & 62.1 & 93.9 & 68.7 & & 83.8 & 97.0 & 92.9 & & 81.2 & 96.7 & 91.4 \\
& & ToD-BERT & & 52.9 & 88.4 & 66.0 & & 83.0 & 91.9 & 92.8 & & 75.8 & 96.7 & 89.6 \\
& & Flan-T5-xl & & 73.8 & 82.2 & 97.0 & & 92.6 & 84.2 & 100.0 & & 99.1 & 96.1 & 99.8 \\
& & +ICL-MU & * & 73.0 & 87.1 & 86.3 & * & 95.0 & 94.6 & 99.1 & * & 98.4 & 98.3 & 99.6 \\
\midrule
10-shot & id-oos & ALBERT & & 27.3 & 87.6 & 42.4 & & 77.8 & 77.6 & 72.2 & & 66.7 & 79.8 & 64.0 \\
& & BERT & & 52.5 & 77.3 & 50.8 & * & 77.5 & 87.5 & 73.8 & & 80.3 & 74.5 & 73.1 \\
& & ELECTRA & & 40.1 & 84.0 & 46.1 & & 79.5 & 85.2 & 75.4 & & 78.0 & 86.5 & 73.3 \\
& & RoBERTa & & 59.7 & 79.1 & 55.8 & & 76.6 & 86.4 & 72.7 & * & 81.0 & 83.9 & 75.8 \\
& & ToD-BERT & & 54.3 & 76.9 & 52.1 & & 80.7 & 79.5 & 75.4 & & 80.6 & 70.2 & 71.9 \\
& & Flan-T5-xl & * & 61.3 & 81.0 & 62.2 & & 69.9 & 75.3 & 67.6 & & 74.3 & 74.9 & 67.3 \\
& & +ICL-MU & & 55.6 & 80.0 & 55.0 & & 79.6 & 77.2 & 74.8 & & 84.2 & 78.9 & 78.2 \\
\midrule
& ood-oos & ALBERT & & 30.5 & 92.7 & 47.1 & & 77.8 & 90.6 & 89.8 & & 66.7 & 95.0 & 85.7 \\
& & BERT & & 64.2 & 91.4 & 68.9 & & 77.5 & 96.8 & 90.0 & & 90.1 & 91.1 & 95.5 \\
& & ELECTRA & & 40.1 & 97.6 & 47.9 & & 79.5 & 94.8 & 90.7 & & 88.6 & 89.1 & 94.2 \\
& & RoBERTa & * & 70.3 & 94.0 & 73.3 & & 89.2 & 95.6 & 95.4 & & 87.5 & 94.6 & 94.0 \\
& & ToD-BERT & & 60.6 & 94.9 & 63.3 & & 86.5 & 96.0 & 94.2 & & 86.5 & 96.4 & 93.7 \\
& & Flan-T5-xl & & 77.2 & 82.7 & 97.2 & & 93.3 & 84.7 & 100.0 & & 99.1 & 95.9 & 99.8 \\
& & +ICL-MU & & 76.9 & 85.9 & 89.8 & * & 95.3 & 93.0 & 99.2 & * & 98.8 & 97.9 & 99.8 \\
\bottomrule
\end{tabular}
}
\caption{Detailed open-world (few-shot) test performance.
    Baselines~\citep{Zhang2021AreDetection} are fine-tuned per task/shot on the training set,
    then use the validation set to tune separate ID-OOS and OOD-OOS thresholds,
    aiming to maximize accuracy on in-scope intent classes (IS-Acc) + OOS recall (OOS-Rcl). 
    In contrast, ICL-MU does not fine-tune parameters (or the OOD-OOS threshold) using target task data.
    For each of the 12 dataset/shot/OOS-type configurations the best model is indicated by a star.
}
\label{tab:results-open-world}
\end{table*}

\subsection{Hand written ICL templates}
\label{appendix:templates}
Here we show the handwritten ICL templates that were used in to create the 
Flan-T5 baselines for intent detection and LEDGAR. 
These were also used to initialize the soft token tags for the initialization 
strategy ``anneal''.
\subsubsection{Intent Detection}
\begin{enumerate}
\item
\begin{description}
\item[icl header] Categorize the following user statements according to their intent.
\item[options header] category options:
\item[demo indicator] example
\item[input indicator] statement:
\item[label indicator] category:
\end{description}
\item
\begin{description}
\item[icl header] Classify the user inquiries below according to their intent.
\item[options header] possible classes:
\item[demo indicator] demonstration
\item[input indicator] inquiry:
\item[label indicator] class:
\end{description}
\item
\begin{description}
\item[icl header] Label these user requests based on their intent type.
\item[options header] label options:
\item[demo indicator] example
\item[input indicator] request:
\item[label indicator] label:
\end{description}
\item
\begin{description}
\item[icl header] Classify these user utterances based on their principal intent.
\item[options header] possible classes:
\item[demo indicator] \#\#\#
\item[input indicator] utterance:
\item[label indicator] class:
\end{description}
\item
\begin{description}
\item[icl header] Determine the intent of the following incoming user requests.
\item[options header] intent options:
\item[demo indicator] e.g.
\item[input indicator] request:
\item[label indicator] intent:
\end{description}
\end{enumerate}

\subsubsection{Legal Text Classification}
\begin{enumerate}
\item
\begin{description}
\item[icl header] Categorize the following contract provisions according to their main topic.
\item[options header] category options:
\item[demo indicator] example
\item[input indicator] provision:
\item[label indicator] category:
\end{description}
\item
\begin{description}
\item[icl header] Classify the contract provisions below according to their main topic.
\item[options header] possible classes:
\item[demo indicator] demonstration
\item[input indicator] provision:
\item[label indicator] class:
\end{description}
\item
\begin{description}
\item[icl header] Label these contract provisions based on their main topic.
\item[options header] label options:
\item[demo indicator] example
\item[input indicator] provision:
\item[label indicator] label:
\end{description}
\item
\begin{description}
\item[icl header] Classify these contract provisions based on their primary topic.
\item[options header] possible classes:
\item[demo indicator] \#\#\#
\item[input indicator] provision:
\item[label indicator] class:
\end{description}
\item
\begin{description}
\item[icl header] Determine the main topic of the following contract provisions.
\item[options header] topic options:
\item[demo indicator] e.g.
\item[input indicator] provision:
\item[label indicator] topic:
\end{description}
\end{enumerate}

\section{LEDGAR Experiments}
To understand whether the learned soft token tags 
have value on tasks outside of intent detection, 
we evaluate them on a version of LEDGAR~\citep{Tuggener2020LEDGAR:Contracts},
a legal text classification dataset.
The objective is to classify provisions in legal contracts.
We start with a version included in LexGLUE~\citep{Chalkidis2021LexGLUE:English}
that has been trimmed down to 100 different classes,
then further remove all provisions longer than 75 tokens.
T5's context window is only 512 tokens,
and we wanted to ensure that at least a few demonstrations fit per test instance.

Because this task is not few-shot, the advantage of using an LLM is limited.
In Table~\ref{tab:ledgar} we see that Flan-T5-XL slightly under-performs
the SBERT (all-mpnet-base-v2) nearest neighbor model.
However, we find that ICL Markup still improves Flan-T5,
pushing its performance beyond this baseline. 

\end{document}